\documentclass{llncs}

\usepackage[utf8]{inputenc}
\usepackage{amsmath}
\usepackage{graphicx}
\usepackage{amsfonts}
\usepackage{amssymb}
\usepackage{url}
\usepackage{array}
\usepackage{float}
\usepackage{multirow}
\usepackage[dvipsnames]{xcolor}

\begin{document}

{\let\thefootnote\relax\footnotetext{Copyright \textcopyright\ 2020 for this paper by its authors. Use permitted under Creative Commons License Attribution 4.0 International (CC BY 4.0). CLEF 2020, 22-25 September 2020, Thessaloniki, Greece.}}

\title{Named entity recognition in chemical patents using ensemble of contextual language models}

\author{Jenny Copara\inst{1,2,3} \and Nona Naderi\inst{1,2} \and Julien Knafou\inst{1,2,3} \and Patrick Ruch\inst{1,2} \and Douglas Teodoro\inst{1,2}}
\institute{University of Applied Sciences and Arts of Western Switzerland, Geneva, Switzerland
\and
Swiss Institute of Bioinformatics, Geneva, Switzerland
\and
University of Geneva, Geneva, Switzerland\\
\email{\{firstname.lastname\}@hesge.ch}
}

\maketitle
\setcounter{footnote}{0} 

\begin{abstract}
Chemical patent documents describe a broad range of applications holding key reaction and compound information, such as chemical structure, reaction formulas, and molecular properties. These informational entities should be first identified in text passages to be utilized in downstream tasks.
Text mining provides means to extract relevant information from chemical patents through information extraction techniques. As part of the Information Extraction task of the Cheminformatics Elsevier Melbourne University challenge, in this work we study the effectiveness of contextualized language models to extract reaction information in chemical patents. We assess transformer architectures trained on a generic and specialised corpora to propose a new ensemble model. Our best model, based on a majority ensemble approach, achieves an exact F$_1$-score of $92.30\%$ and a relaxed F$_1$-score of $96.24\%$. The results show that ensemble of contextualized language models can provide an effective method to extract information from chemical patents.
\end{abstract}

\section{Introduction}

Chemical patents represent a valuable information resource in downstream innovation applications, such as drug discovery and novelty checking. However, the discovery of chemical compounds described in patents is delayed by a few years \cite{He2020a}. Among the reasons, it could be considered the complexity of the chemical patent information sources \cite{Habibi2016}, the recent increase in the number of chemical patents without manual curation, and the particular wording used in the domain. Narratives in chemical patents contain often concepts expressed in a way to protect or hide information, as opposed to scientific literature, for example, where the text tends to be as clear as possible~\cite{Valentinuzzi2017}. In this landscape, information extraction methods, such as Named Entity Recognition (NER), provide a suited solution to identify key information in patents. 

NER aims to identify information of interest and their respective instances in a document~\cite{Grishman2019,Okurowski1993}.
It has been often addressed as a sequence classification task, where a sequence of features, usually tokens, is used to predict the class of a text passage. One of the most successful approaches in sequence classification is Conditional Random Fields (CRF) \cite{Lafferty2001,Sutton2012}. CRF was proposed to solve sequence classification problems by estimating the conditional probability of a label sequence given a word sequence, considering a set of observed features in the latter.  It was established as the state-of-the-art in different NER domains for many years \cite{Leaman2008,Rocktschel2012,Leaman2015,Ratinov2009,Guo2014,Habibi2016,Yadav2018}. In the chemical patent domain, CRF was explored by Zhang \emph{et al.}~\cite{Zhang2016} in the CHEMDNER patent corpus~\cite{Krallinger2015}. Using a set of hand-crafted and unsupervised features derived from word embeddings and Brown clustering, their model achieved $87.22\%$ of F$_1$-score. With similar F$_1$-score performance, Akhondi \emph{et al.} 
~\cite{Akhondi2016} explored CRF combined with dictionaries in the biomedical domain in the tmChem tool \cite{Leaman2015} in order to select the best vocabulary for the CHEMDNER patent corpus. It has been shown \cite{Habibi2016} that recognizing chemical entities in the full patent text is a harder task than in titles and abstracts, due the peculiarities of the chemical patent text. Evaluation in full patents was performed using BioSemantics patent corpus
~\cite{Akhondi2014} through neural approaches based on the Bidirectional Long-Short Term Memory (BiLSTM) CRF \cite{Habibi2017} and the BiLSTM Convolutional Neural Network (CNN) CRF \cite{Zhai2019} architectures, with performance of $82.01\%$ and $85.68\%$ of F$_1$-score, respectively. It is worth noting that for the first architecture \cite{Habibi2017}, the authors used word2vec embeddings~\cite{Mikolov2013} to represent features, while in the latter \cite{Zhai2019}, the authors used ELMo contextualized embeddings \cite{Peters2018}.

Over the years, neural language models have improved their ability to encode the semantics of words using large amounts of unlabeled text for self-supervised training. They have initially evolved from a straightforward model \cite{Bengio2003} of one hidden layer that predicts the next word in a sequence, aiming to learn the distributed representation of words (i.e., the word embedding vector), to an improved objective function that allows learning from larger amounts of text \cite{Collobert2011}, using higher computational resources and with longer training time. These developments have encouraged the seeking of language models able to bring high-quality word embeddings with lower computational cost (i.e., word2vec~\cite{Mikolov2013} and Global Vectors (GloVe)~\cite{Pennington2014}). However, natural language still presented challenges for language models, in particular, concerning word contexts and homonyms. More recently, a second type of word embeddings have attracted attention in the literature, the so-called contextualized embeddings, such as ELMo, UMLFiT~\cite{Howard2018}, GPT-2~\cite{Radford2019}, and BERT~\cite{Devlin2019}. Particularly, the BERT architecture uses the attention mechanism to train deep bidirectional token representations, conditioning tokens on their left and right contexts.

In this work, we explore contextualized language models to extract information in chemical patents as part of the Named Entity Recognition task of the Information extraction from Chemical Patents (ChEMU) lab ~\cite{He2020a,He2020b}. Pre-trained contextualized languages models, based on the BERT-based architecture, are used as baseline model and fine-tuned on the examples of the ChEMU NER task to classify tokens according to the different entities. In the challenge, the corpus was annotated with the entities: \textit{example\_label}, \textit{other\_compound}, \textit{reaction\_product}, \textit{reagent\_catalyst}, \textit{solvent}, \textit{starting\_material}, \textit{temperature}, \textit{time}, \textit{yield\_other}, and \textit{yield\_percent}. We investigate the combination of different architectures to improve NER performance. In the following sections, we describe the design and results of our experiments.

\section{Methods and data}

\subsection{NER model}
\subsubsection{Transformers with a token classification on top.}

We assess five language models based on the transformers architecture to classify tokens according to the named-entities classes. The first four models are variations of the BERT model in terms of size and tokenization: \emph{bert-base-cased}, \emph{bert-base-uncased}, \emph{bert-large-cased}, and \emph{bert-large-uncased}. These models were originally pretrained on a large corpus of English text extracted from BookCorpus~\cite{Zhu2015} and Wikipedia, with different number of attention heads for the base and large types (12 and 16 respectively). The fifth pretrained language model assessed is ChemBERTa\footnote{\url{https://github.com/seyonechithrananda/bert-loves-chemistry}}, a RoBERTa-based transformer architecture \cite{Liu2019}, trained on a corpus of 100k Simplified Molecular Input Line Entry System (SMILES)~\cite{Weininger1988} strings from the ZINC benchmark dataset~\cite{Irwin2005}.

 Our models consist of BERT models specialised for NER, with a fully connected layer on top of the hidden states of each token. They are fine-tuned on the ChEMU Task 1 dataset, using the train and development sets provided. The fine-tuning is performed with a sequence length of 256 tokens, a warmup proportion of $0.1$ (percentage of warmup steps with respect to the total amount of steps), and a batch size of 32. 
The tokenization process is driven by the original model's tokenizer, i.e., for the BERT-based models, WordPiece~\cite{Wu2016} is applied, while for the RoBERTa-based model, Byte-Pair-Encoding~\cite{Sennrich2016} is applied. The Adam optimizer is employed to optimize network weights~\cite{Kingma2015}. The first four language models are fine-tuned for 10 epochs and a learning rate of $3e-5$. For ChemBERTa model, we conduct a grid search over the development set and found the best performance around 29 epochs of fine-tuning and a learning rate of $4e-5$. The implementations are based on the Huggingface framework.\setcounter{footnote}{1}  \footnote{\url{https://huggingface.co/transformers/}} 

\subsubsection{Ensemble model.}

Our ensemble method is based on a voting strategy, where each model votes with its predictions and a simple majority of votes is necessary to assign the predictions~\cite{Copara2020}. In other words, for a given document, our models infer their predictions independently for each entity, then, a set of passages that received at least a vote is taken into consideration for casting votes. This means that, for a given document and a given entity, we end up with multiple passages associated with a number of votes, then, again for a given entity, the ensemble method will predict as positive all the passages that get the majority of votes. Note that each entity is predicted independently and that the voting strategy does allow the fact that a passage could have been labeled as positive for multiple entities at once. 

 Finally, in order to decide on the optimal composition of the ensemble model, we used the development set and compute all possible ensemble predictions using the above methodology. As we had 7 models in total, we tried every possible combination from 2 to 7 models. We retained the ensemble composition with the best overall F$_1$-score and used it for the test set. Originally, the ensemble model giving the best F$_1$-score was combining \emph{bert-large-uncased}, \emph{bert-base-cased}, CRF, \emph{bert-base-uncased} and the CNN model (5 models). However, due to the size of the test set (approximately 10k patent snippets), we had to discard the large models of the ensemble strategy due to their much higher algorithmic complexity and the time constraints. The retained models in the ensemble were then \emph{bert-base-cased}, \emph{bert-base-uncased} and the CNN model.

\subsubsection{Baseline.}

We consider two models for our baseline: CRF and CNN. For the CRF model, a set of standard features in a window of $\pm 2$ tokens are created without taking into account part-of-speech tags, neither gazetteers. The features used are token itself, lower-cased word, capitalization pattern, type of token (i.e., digit, symbol, word), 1-4 character prefixes/suffixes, digit size (i.e., size 2 or 4), combination of values (digit with alphanumeric, hyphen, comma, period), binary features for upper/lower-cased letter, alphabet/digit char and symbol. Please refer to \cite{Copara2016,Guo2014} for further details on the features used. The CRF classifier implementation relies on the \textit{CRFSuite}.\footnote{\url{http://www.chokkan.org/software/crfsuite/}}

The CNN model~\cite{Lecun1989} for NER relies on incremental parsing with Bloom embeddings, a compression technique for neural network models dealing with sparse high-dimensional binary-coded instances~\cite{Serr2017}. The convolutional layers use residual connections, layer normalization and maxout non-linearity. The input sequence is embedded in a vector compounded by Bloom embeddings modeling the characters, prefix, suffix and part-of-speech of each word. Convolutional filters of 1D are used over the text to predict how the next words are going to change. Our implementation relies on the spaCy NER module,~\footnote{\url{https://spacy.io}} using the pretrained transformer \emph{bert-base-uncased} for 30 epochs and a batch size of 4. During the test phase, we fixed the max size of the text to 1.5M due to RAM memory limitations.

\subsection{Data}

\begin{figure}
\centering
\includegraphics[width=\textwidth]{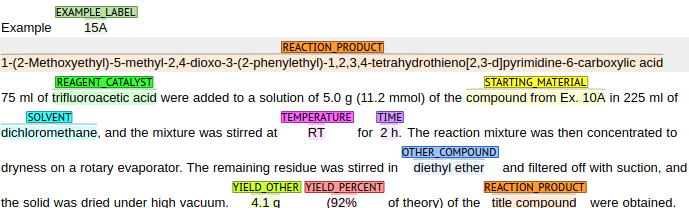}
\caption{Data example with annotations for the ChEMU NER task.} 
\label{figure:example-data}
\end{figure}

The data in ChEMU Task 1 (NER) is provided as snippets sampled from 170 English patents from the European Patent Office and the United States Patent and Trademark Office \cite{He2020a,He2020b}. Gold annotations were provided for training (900 snippets) and development (250 snippets) sets for a total of $20,186$ entities. The annotation was done in the BRAT standoff format. Fig. \ref{figure:example-data} shows an example of a snippet with annotations for several entities, including \emph{reaction\_product} (two annotations), \emph{starting\_material} and \emph{temperature}. 

During the development phase, we used the official development set as our test set. The official training set was split into train and development sets in order to train the weights and tune hyper parameters of our models, respectively. As a result of this new setting, 800 snippets were available in train set, 100 in the development set and 225 in test set. Table \ref{table:data} shows the entity distribution during the development phase. The majority of the annotations come from \textit{other\_compound}, \textit{reaction\_product} and \textit{starting\_material}, covering the 52\% of entities in the development phase. In contrast, \textit{example\_label}, \textit{time} and \textit{yield\_percent} entities represent $17\%$ of entities in the development phase.

\begin{table}
\centering
\caption{Entity distribution in the development phase based on the official training and development sets. Test set is the official development set. Dev set is random set extracted from the official training set.}
\label{table:data}
\begin{tabular}{|l|l|l|l|l|}
\hline
\textbf{Entity} & \begin{tabular}[c]{@{}l@{}}\textbf{Train} \\ (count/\%)\end{tabular} & \begin{tabular}[c]{@{}l@{}}\textbf{Dev} \\ (count/\%)\end{tabular} & \begin{tabular}[c]{@{}l@{}}\textbf{Test} \\ (count/\%)\end{tabular} & \begin{tabular}[c]{@{}l@{}}\textbf{All} \\ (count/\%)\end{tabular} \\ \hline
example\_label & 784/5 & 102/5 & 218/6 & 1104/5 \\ \hline
other\_compound & 4095/28 & 545/29 & 1080/28 & 5720/28 \\ \hline
reaction\_product & 1816/13 & 236/12 & 506/13 & 2558/13 \\ \hline
reagent\_catalyst & 1135/8 & 146/8 & 289/8 & 1570/8 \\ \hline
solvent & 1001/7 & 139/7 & 250/7 & 1390/7 \\ \hline
starting\_material & 1543/11 & 211/11 & 413/11 & 2167/11 \\ \hline
temperature & 1345/9 & 170/9 & 346/9 & 1861/9 \\ \hline
time & 928/6 & 131/7 & 252/7 & 1311/6 \\ \hline
yield\_other & 940/7 & 121/6 & 261/7 & 1322/7 \\ \hline
yield\_percent & 848/6 & 107/6 & 228/6 & 1183/6 \\ \hline
All & 14435/100 & 1908/100 & 3843/100 & 20186/100 \\ \hline
\end{tabular}
\end{table}

\subsection{Evaluation metrics}

The metrics used to evaluate the models are precision, recall, and F$_1$-score. As it can be seen in the example of Fig. \ref{figure:example-data}, each entity has a span that is expected to be identified by the NER models as well as the correct entity type. The evaluation for the challenge is established under strict and relaxed span matching conditions~\cite{He2020a,He2020b}. The exact matching condition takes into account the correct identification of both, span and entity type. On the other hand, the relaxed matching condition evaluates how accurate is the predicted span concerning the real. Our models are evaluated with the ChEMU web page system for the official results~\footnote{\url{http://chemu.eng.unimelb.edu.au/}} and with the BRAT Eval tool for the offline analyses~\footnote{\url{https://bitbucket.org/nicta_biomed/brateval/src/master/}}. 

\section{Results and discussion}

In this section, we present the results of our models in the development and official test phases. Additionally, we perform error analyses on the results of the test set used in the development phase for some relevant models.

\subsection{Model's performance in the development phase}

Table \ref{table:fscores-dev} shows the exact and relaxed overall F$_1$-scores for all the models explored by our team in the development phase of the ChEMU NER task. As we can see, the ensemble model outperforms all the individual models for both exact and relaxed metrics. On the other hand, despite being trained on a specialised corpus, ChemBERTa achieves the lowest performance. The reported results come from the ChEMU official evaluation web page except for the CNN, \emph{bert-large-uncased}, and the ensemble models, which are provided by the BRAT Eval tool.

\begin{table}
\centering
\caption{Performance of the different models in the development phase in terms of F$_1$-score. *models evaluated using the BRAT Eval tool.}
\label{table:fscores-dev}
\begin{tabular}{|l|c|c|c|c|c|c|c|c|}
\hline
\textbf{Metric} & \textbf{CRF} & \textbf{CNN*} & \multicolumn{2}{c|}{\textbf{bert-base}}  & \multicolumn{2}{c|}{\textbf{bert-large}}  & \textbf{Chem}  & \textbf{Ensemble*} \\ \cline{4-7}

 &  &  & \textbf{cased} & \textbf{uncased} & \textbf{cased} & \textbf{uncased*} & \textbf{BERTa} &  \\ \hline
exact & 0.8722 & 0.8182 & 0.9140 & 0.9113 & 0.9079 & 0.9052 & 0.6810 & \textbf{0.9285} \\ \hline
relaxed & 0.9450 & 0.8820 & 0.9732 & 0.9719 & 0.9706 & 0.9910 & 0.8500 & \textbf{0.9876} \\ \hline
\end{tabular}
\end{table}

The results of all models with respect to the individual entities are presented in Table~\ref{table:results-dev-exact}. As for the overall results, the ensemble model outperforms the individual models for all entities apart from \textit{time}, for which the bert-base-cased presents the best performance. The highest improvement for the ensemble model is seen for the \textit{reaction\_product} and \textit{starting\_material} entities with over 12-point increase in F$_1$-score. Considering only the individual models, the bert-base models outperform the other individual models, including the bert-large models, for all the entities, apart from \textit{starting\_material}, for which the CNN model has the best performance.

\begin{table}
\caption{Evaluation results on the development set for the exact F$_1$-score metric.}
\label{table:results-dev-exact}
\centering 
\begin{tabular}{|l|c|c|c|c|c|c|c|c|}
\hline
\bf{Entity}& \textbf{CRF} & \textbf{CNN} & \multicolumn{2}{c|}{\textbf{bert-base}} & \multicolumn{2}{c|}{\textbf{bert-large}}  & \textbf{Chem}  & \textbf{Ensemble} \\ \cline{4-7}
 &  &  & \textbf{cased} & \textbf{uncased} & \textbf{cased} & \textbf{uncased} & \textbf{BERTa} &  \\ \hline
example\_label& 0.9630 & 0.9526 & 0.9862 & 0.9817  &0.9793 & 0.9769 &0.9631 &\bf{0.9885}\\ \hline
other\_compound&  0.8762 &0.7409 & 0.8953 & 0.8938& 0.8947& 0.8925 &0.7850&\bf{0.9052}\\ \hline
reaction\_product& 0.7535 & 0.8425 & 0.8586 &0.8515 &  0.8410& 0.8427  & 0.5957&\bf{0.8807}\\ \hline
reagent\_catalyst& 0.8330 & 0.8557 & 0.8595  & 0.8355 & 0.8498 & 0.8468 &  0.4673&\bf{0.8946}\\ \hline
solvent& 0.8949 & 0.7517 & 0.9447 & 0.9451 & 0.9407 &0.9426  & 0.5945&\bf{0.9545}\\ \hline
starting\_material&0.7253  & 0.8229 &0.8072 &0.8153  &0.7995  &0.7813  & 0.4405&\bf{0.8470} \\ \hline
temperature&0.9796 & 0.6397 & 0.9842 & 0.9842 & 0.9827 & 0.9841 & 0.8105&\bf{0.9855} \\ \hline
time& 0.9900 & 0.8533 & \bf{1.0000} & 0.9941 & 0.9941 &0.9941  & 0.8141 & 0.9980\\ \hline
yield\_other&0.9046 &0.9448  &0.9905  &  0.9924 & 0.9811 & 0.9848 &0.7135 &\bf{0.9943} \\ \hline
yield\_percent& 0.9913 &0.9693  &\bf{0.9978}  & \bf{0.9978} & 0.9913 & 0.9892 & 0.7131&\bf{0.9978} \\ \hline
\end{tabular}
\end{table}

The ensemble model achieves the best performance for the \textit{time}, \textit{yield\_other} and \textit{yield\_percent} entities. We believe this is due to the patterns observed for them in the training and test data. For example, for the \textit{yield\_percent} entity, the pattern is mostly a number followed by the percentage symbol (`\%'). Similarly, for the \textit{time} entity, the instances usually appear as a number followed for a time-indicator word. On the other hand, the \textit{reaction\_product}, \textit{reagent\_catalyst} and \textit{starting\_material} entities show the lowest performance, with $88.07\%$, $89.46\%$ and $84.70\%$ of F$_1$-score, respectively. These entities are of chemical types, often molecule strings (e.g.,  4-(6-Bromo-3-methoxypyridin-2-yl)-6-chloropyrimidin-2-amine)~\cite{He2020a,He2020b}. As our models did not include a post-processing step, as proposed in~\cite{teodoro2010automatic}, these entities were sometimes recognized partially as a result of the language model sub-word tokenization process.

During the development phase, we also investigate the performance of ChemBERTa. As ChemBERTa is a language model trained on the chemical domain, it is expected to achieve competitive results. However, for the NER downstream task in chemical patents, the results go in a different direction. As shown in 
~Table \ref{table:results-dev-exact}, ChemBERTa obtains the lowest results among all the explored models for both exact and relaxed metrics. We believe that the size of the corpus used to train the other explored language models has led to better chemical entity representations. Additionally, as the task aims to identify other entities than molecules, the ChemBERTa model naturally fails as its train set is only based on SMILES strings.

\subsection{Model's performance in the test phase}

In the official test phase, $9,999$ files containing snippets from chemical patents were available for evaluating the models. We submitted 3 official runs: run 1, based on the baseline CRF model; run 2, based on the bert-base-cased model; and run 3, based on the ensemble model. Table \ref{table:scores-test} shows the official performance of our models for the exact and relaxed span matching metrics in terms of F$_1$-score. The ensemble model achieves $92.30\%$ of exact F$_1$-score, yielding more than 11-point improvement over our baseline and at least 1-point improvement over the best individual contextualized language model (bert-base-cased). It outperforms run 1 and run 2 for all the entities in both exact and relaxed metrics. We believe that the performance difference between the CRF model and the ensemble model is due mostly to the fact that language models based on attention mechanisms are able to provide better contextual feature representations without the specific design of hand-crafted features as in the case of CRF. 

\begin{table}
\centering
\caption{Official performance of our models in terms of F$_1$-score for the exact and relaxed metrics.}
\label{table:scores-test}
\begin{tabular}{|l|c|c|c|c|c|c|}
\hline
\multirow{2}{*}{\textbf{Entity}} & \multicolumn{2}{c|}{\textbf{CRF}} & \multicolumn{2}{c|}{\textbf{bert-base-cased}} & \multicolumn{2}{c|}{\textbf{Ensemble}} \\ \cline{2-7} 
 & \textbf{exact} & \textbf{relaxed} & \textbf{exact} & \textbf{relaxed} & \textbf{exact} & \textbf{relaxed} \\ \hline
example\_label & 0.9190 & 0.9367 & 0.9617 & 0.9730 & 0.9669 & 0.9784 \\ \hline
other\_compound & 0.8310 & 0.9029 & 0.8780 & 0.9608 & 0.8920 & 0.9653 \\ \hline
reaction\_product & 0.6462 & 0.7689 & 0.8593 & 0.9378 & 0.8766 & 0.9322 \\ \hline
reagent\_catalyst & 0.7598 & 0.8035 & 0.8791 & 0.9082 & 0.9022 & 0.9176 \\ \hline
solvent & 0.8299 & 0.8323 & 0.9444 & 0.9491 & 0.9541 & 0.9541 \\ \hline
starting\_material & 0.4957 & 0.6752 & 0.8413 & 0.9343 & 0.8701 & 0.9394 \\ \hline
temperature & 0.9499 & 0.9688 & 0.9692 & 0.9902 & 0.9729 & 0.9877 \\ \hline
time & 0.9698 & 0.9843 & 0.9868 & 0.9967 & 0.9879 & 0.9978 \\ \hline
yield\_other & 0.8984 & 0.8984 & 0.9799 & 0.9821 & 0.9842 & 0.9865 \\ \hline
yield\_percent & 0.9705 & 0.9807 & 0.9936 & 0.9962 & 0.9974 & 0.9974 \\ \hline
ALL & 0.8056 & 0.8683 & 0.9098 & 0.9596 & 0.9230 & 0.9624 \\ \hline
\end{tabular}
\end{table}

The 5-top best performing entities identified by our models are \textit{example\_label}, \textit{temperature}, \textit{time}, \textit{yield\_other}, \textit{yield\_percent}, which is similar to the results found in the development phase. For all of our submissions, the entity with lowest performance in the official test phase is \textit{starting\_material}, achieving $49.57\%$, $84.13\%$ and $87.01\%$ of exact F$_1$-score in the CRF, bert-base-cased and ensemble models, respectively. As we will see further in the error analyses section, this entity is often confused with the \emph{reagent\_catalyst} entity in the development phase. From the chemistry point of view, both starting material (reactants) and catalysts (reagents) entities are present at the start of the reaction, with the difference that the latter is not consumed by the reaction. These terms are often used interchangeably though, which could be the reason for the confusion. Despite the much larger size of the test set (approximately 10 times the size of the training set), these results suggest that the test set has a similar entity distribution of the dataset provided in the development phase. 

In Table \ref{tab:rank-submissions} is shown a summary of the top ten official results, including our runs 2 and 3 (BiTeM team, ranked 6 and 7), the best model and the challenge baseline. If we consider the exact F$_1$-score metric, our ensemble model shows at least 3-point improvement from the ChEMU Task 1 NER baseline and more than 3-point behind the top 1. For the relaxed metric, our best model performs slightly better, showing more than 5-point improvement from the baseline and less than 1-point below the top system. 

\begin{table}
\centering
\caption{Official BiTeM results compared to the best model and the BANNER baseline.}
\label{tab:rank-submissions}
\begin{tabular}{|c|l|c|c|c|c|c|c|}
\hline
\multirow{2}{*}{\textbf{Rank}} & \multirow{2}{*}{\textbf{Team}} & \multicolumn{2}{c|}{\textbf{Precision}} & \multicolumn{2}{c|}{\textbf{Recall}} & \multicolumn{2}{c|}{\textbf{F$_1$-score}} \\ \cline{3-8}
  &  & \textbf{exact} & \textbf{relaxed} & \textbf{exact} & \textbf{relaxed} & \textbf{exact} & \textbf{relaxed} \\ \cline{1-8}
1 &  Melaxtech & 0.9571 & 0.9690 & 0.9570 & 0.9687 & 0.9570 & 0.9688 \\ \hline
6 &  \bf{BiTeM (run 3)} & 0.9378 & 0.9692 & 0.9087 & 0.9558 & 0.9230 & 0.9624 \\ \hline
7 &  \bf{BiTeM (run 2)} & 0.9083 & 0.9510 & 0.9114 & 0.9684 & 0.9098 & 0.9596 \\ \hline
10 &  Baseline (BANNER) & 0.9071 & 0.9219 & 0.8723 & 0.8893 & 0.8893 & 0.9053 \\ \hline
\end{tabular}
\end{table}

The performance of the ensemble model for all entities on test set in terms of precision, recall and F$_1$-score for both exact match and relax is presented in Table~\ref{tab:ensemble-p-r}. The best precision and recall for the exact match metric are achieved for the \textit{yield\_percent} entity, reaching 99.74\% and 99.74\%, respectively. Overall, precision is always above 93\% for the relaxed metric and at least 88\% for the exact metric.

\begin{table}
\centering
\caption{Performance of the ensemble model for all entities on the test set in terms of precision, recall and F$_1$-score}
\label{tab:ensemble-p-r}
\begin{tabular}{|l|c|c|c|c|c|c|}
\hline
\multirow{2}{*}{\bf{Entity}}    &\multicolumn{2}{c|}{\bf{Precision}}    &\multicolumn{2}{c|}{\bf{Recall}} &\multicolumn{2}{c|}{\bf{F$_1$-score}} \\\cline{2-7}
    &\bf{exact} &\bf{relaxed}    &\bf{exact} &\bf{relaxed} &\bf{exact} &\bf{relaxed} \\\hline
example\_label  &0.9711     &0.9827     &0.9628     &0.9742 &   0.9669  & 0.9784\\\hline
other\_compound &0.9197     &0.9730     &0.8659     &0.9578 &   0.8920  & 0.9653\\\hline
reaction\_product&0.8942    &0.9367     &0.8596     &0.9277 &   0.8766  & 0.9322 \\\hline
reagent\_catalyst&0.9268    &0.9435     &0.8790     &0.8931 &   0.9023	& 0.9176\\\hline
solvent         &0.9620     &0.9620     &0.9463     &0.9463 &   0.9541	& 0.9541\\\hline
starting\_material&0.8886   &0.9545     &0.8523     &0.9247 &   0.8701	& 0.9394\\\hline
temperature     &0.9769     &0.9901     &0.9690     &0.9852 &   0.9729	& 0.9876\\\hline
time            &0.9846     &0.9956     &0.9912     &1.0000 &   0.9879	& 0.9978\\\hline
yield\_other    &0.9776     &0.9798     &0.9909     &0.9932 &   0.9842	& 0.9865\\\hline
yield\_percent  &0.9974     &0.9974     &0.9974     &0.9974 &   0.9974	& 0.9974\\\hline
\end{tabular}
\end{table}

Lastly, our CRF baseline achieves $80.56\%$ of exact F$_1$-score, while the competition baseline, which is based also on CRF, but customized for biomedical NER, taking into account features, such as part-of-speech, lemma, Roman numerals, names of the Greek letters, achieves $88.93\%$~\cite{Leaman2008}. Indeed, we believe those features give the advantage to the competition baseline as they could better characterize chemical entities.

\subsection{Error analysis}

\begin{figure}
    \centerline{\includegraphics[scale=0.26]{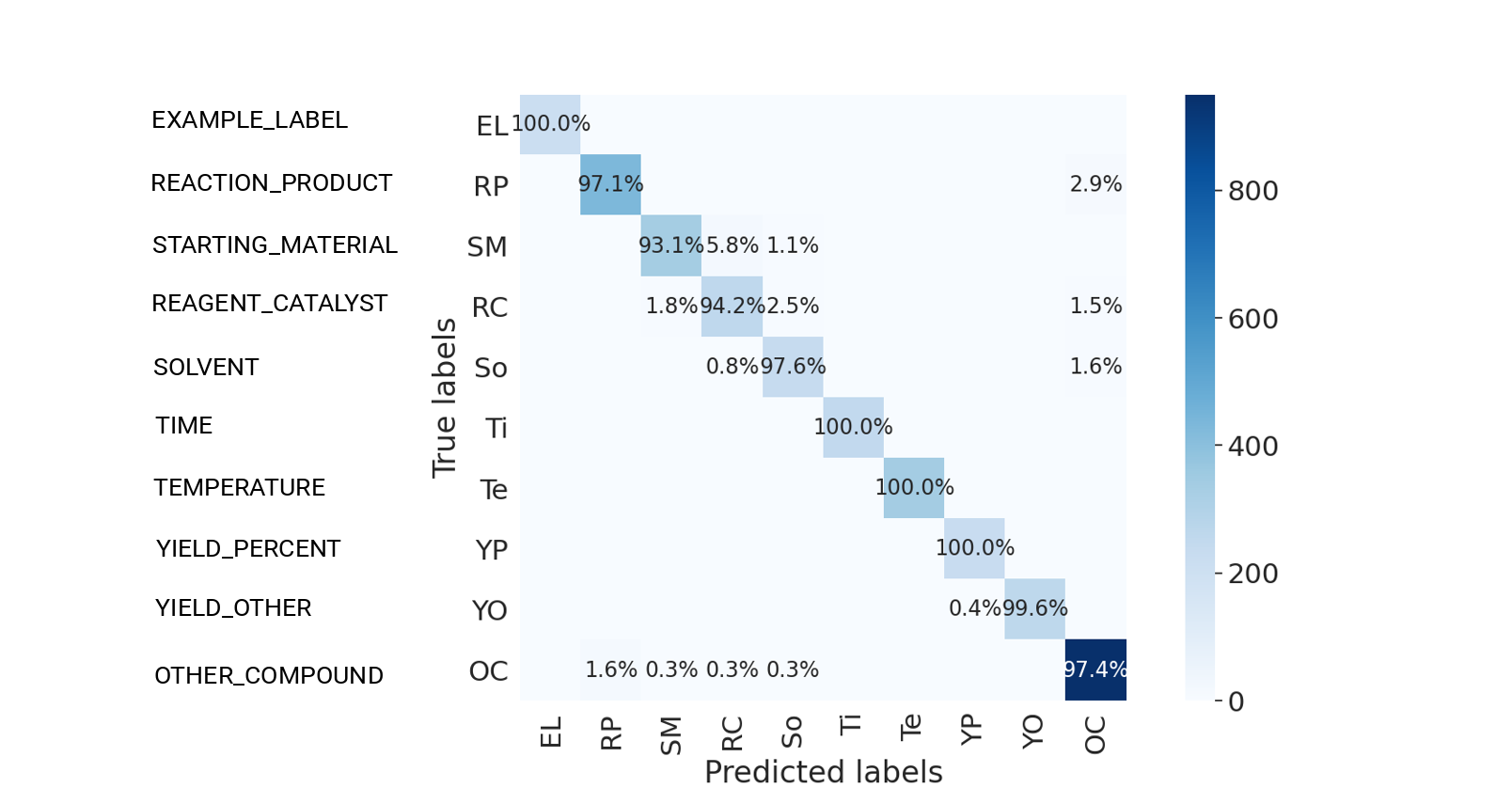}}
        \caption{Normalized confusion matrix for the ensemble model predictions on the official development set. Only exact matches are considered.}
    \label{fig:confusion_matrix}
\end{figure}

As the gold annotations for the test set are not available, we perform the error analysis on the official development set (used as our test set in the development phase, see Table~\ref{table:data}). Fig.\ \ref{fig:confusion_matrix} shows the confusion matrix for the ensemble predictions for the exact metric. As we can see, most confusion occurred for the \textit{starting\_material} entity, which is mostly confused with \textit{reagent\_catalyst}, and for the \textit{reaction\_product} entity, which is mistaken for \textit{other\_compound}. As mentioned previously, these entities - material/reactant and catalyst/reagent, and product/compound - are often used interchangeably in chemistry passages, which is likely the reason for the model's confusion.

The error analysis of the incorrectly identified spans by the ensemble model shows that in almost 78.8\% of the cases, the predicted entity was longer in length, for example, \textit{sodium thiosulfate aqueous} instead of \textit{aqueous} and \textit{concentrated hydrochloric acid} instead of \textit{hydrochloric acid}. The entities that are partially detected are mainly \textit{starting\_material}, which is inconsistently annotated in some cases, as \textit{Intermediate 13/6/21} (predicted as \textit{13/6/21} by the ensemble model), and in some cases as only the number, such as \textit{3} (predicted as \textit{Intermediate 3} by the ensemble model). 42.3\% of the span errors were multi-word entities.



\begin{figure}
    \centering
    \includegraphics[scale=0.55]{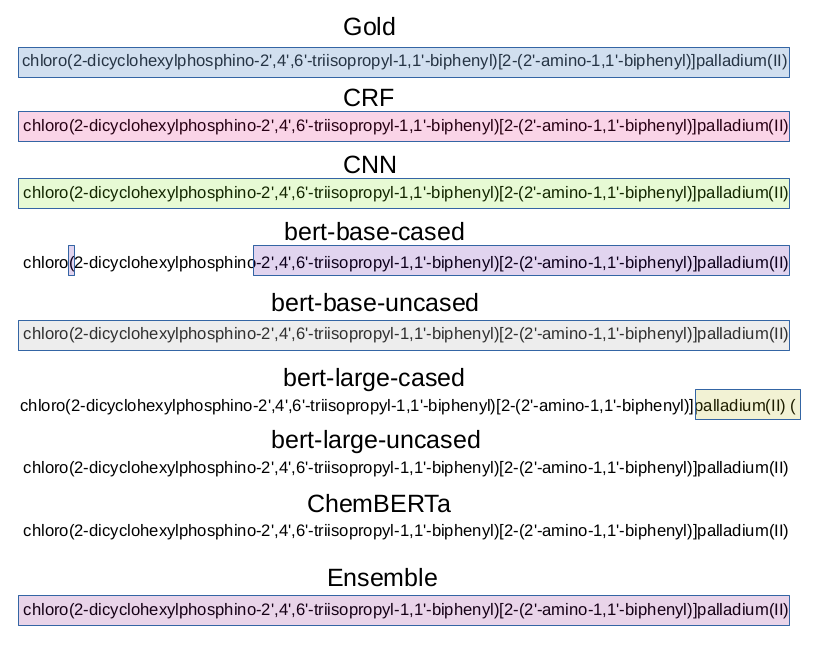}
 \caption{An example of predictions by different models for (reagent\_catalyst) annotation. The span detected by each model is color-coded.}
    \label{fig:prediction_example}
\end{figure}

Fig.\ \ref{fig:prediction_example} shows how different models detected a reagent catalyst entity described by a long text span. It seems that entities with longer text span, such as \textit{reagent\_catalyst}, \textit{other\_compound}, \textit{reaction\_product}, and \textit{starting\_material}, are less likely to be correctly detected by the contextualized language models. The bert-large-uncased and ChemBERTa models did not detect any token of the entity while both bert-large-cased and bert-base-cased models were able to only partially detect the entity. Particularly, the larger nature of the BERT large models was not translated into more effective representations for these entities.  

\begin{figure}
\centering
\includegraphics[scale=0.6]{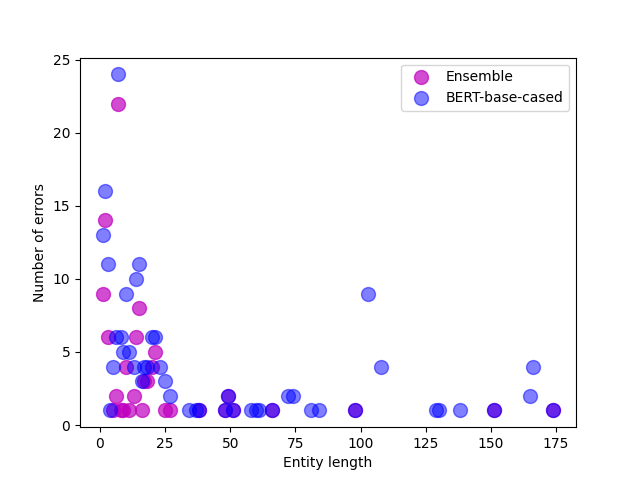}
\caption{Number of span errors by the ensemble and BERT-base-cased models based on the length of the entities (in character).}
\label{fig:spanerrors}
\end{figure}

Figure~\ref{fig:spanerrors} shows the comparison of the span errors of the ensemble and BERT-base-cased models based on the length of entities (in character). While most errors of both models are focused on smaller entities, the BERT-base-cased model makes more mistakes than the ensemble model in detecting the spans of the longer entities. We believe this effect could be also related to the sub-word tokenization process of transformers. The combination of models smooths the effect in the ensemble model. 

\section{Conclusions}

In this task, we explored the use of contextualized language models based on the transformer architecture to extract information from chemical patents. The combination of language models resulted in an effective approach, outperforming the baseline CRF model but also individual transformer models. Our experiments show that without extensive pre-training in the patent chemical domain, the majority vote approach is able to leverage distinctive features present in the English language, achieving $92.30\%$ of exact F$_1$-score in the ChEMU NER task. It seems that the transformer models are able to take advantage of natural language contexts in order to capture the most relevant features without supervision in the chemical domain. Our next step will be to investigate pre-trained models on large chemical patent corpora to further improve the NER performance.

\bibliographystyle{splncs04}
\bibliography{references}

\end{document}